\documentclass[
]{ceurart}

\usepackage{listings}
\begin{document}

\copyrightyear{2023}
\copyrightclause{Copyright for this paper by its authors.
  Use permitted under Creative Commons License Attribution 4.0
  International (CC BY 4.0).}

\conference{CAMLIS'23: Conference on Applied Machine Learning for Information Security,
  October 19--20, 2023, Arlington, VA}

\title{Multi-Agent Reinforcement Learning for Maritime Operational Technology Cyber Security}


\author[1]{Alec Wilson}[%
email=Alec.Wilson@uk.bmt.org,
orcid=0009-0003-9181-9766,
]

\address[1]{BMT, London, UK}


\author[1]{Ryan Menzies}[%
email=Ryan.Menzies@uk.bmt.org,
orcid=0009-0003-9646-196X,
]

\author[1]{Neela Morarji}[%
orcid=0000-0001-9002-1187,
]

\author[2]{David Foster}[%
]
\address[2]{ADSP, London, UK}

\author[1]{Marco Casassa{\space}Mont}[%
orcid=0009-0004-7611-6947,
]

\author[1]{Esin Turkbeyler}[
]

\author[1]{Lisa Gralewski}[%
]


\begin{abstract}
  This paper demonstrates the potential for autonomous cyber defence to be applied on industrial control systems and provides a baseline environment to further explore Multi-Agent Reinforcement Learning's (MARL) application to this problem domain.  It introduces a simulation environment, IPMSRL, of a generic Integrated Platform Management System (IPMS) and explores the use of MARL for autonomous cyber defence decision-making on generic maritime based IPMS Operational Technology (OT).
  
OT cyber defensive actions are less mature than they are for Enterprise IT. This is due to the relatively ‘brittle’ nature of OT infrastructure originating from the use of legacy systems, design-time engineering assumptions, and lack of full-scale modern security controls. There are many obstacles to be tackled across the cyber landscape due to continually increasing cyber-attack sophistication and the limitations of traditional IT-centric cyber defence solutions. Traditional IT controls are rarely deployed on OT infrastructure, and where they are, some threats aren’t fully addressed.

In our experiments, a shared critic implementation of Multi Agent Proximal Policy Optimisation (MAPPO) outperformed Independent Proximal Policy Optimisation (IPPO). MAPPO reached an optimal policy (episode outcome mean of 1) after 800K timesteps, whereas IPPO was only able to reach an episode outcome mean of 0.966 after one million timesteps. Hyperparameter tuning greatly improved training performance. Across one million timesteps the tuned hyperparameters reached an optimal policy whereas the default hyperparameters only managed to win sporadically, with most simulations resulting in a draw. We tested a real-world constraint, attack detection alert success, and found that when alert success probability is reduced to 0.75 or 0.9, the MARL defenders were still able to win in over 97.5\% or 99.5\% of episodes, respectively.

\end{abstract}


\maketitle

\section{Introduction}

Operational Technology (OT) cyber defensive actions are less mature than they are for Enterprise IT. This is due to the relative ‘brittle’ nature of OT infrastructure originating from the use of legacy systems, design-time engineering assumptions, and lack of full-scale modern security controls. Traditional IT controls are rarely deployed on OT infrastructure, and where they are, some threats aren’t fully addressed. Additionally, there may be a lack of trained cyber personnel available in deployed operational OT environments e.g., at-sea vessels, hence the opportunity to develop autonomous defensive systems.
Reinforcement learning (RL)  \cite{barto_learning_nodate} \cite{sutton_reinforcement_2018} is a subset of machine learning that allows an AI-driven system (sometimes referred to as an agent) to learn through trial and error using feedback signals (through rewards) from its actions. An agent executes autonomous actions and is given a reward (or punishment) contingent with the consequences of these actions in an environment. The agent adapts its strategy, referred to as a policy, to maximize cumulative reward.

RL has seen key recent developments in various domains such as game environments e.g., Go \cite{silver_mastering_2016} and real world problems e.g., video compression \cite{mandhane_muzero_2022}. Many domains, including cyber-security, are now using RL to train within simulated environments. There are a few examples of cyber focused RL environments in the literature, notably: Yawning Titan \cite{andrew_developing_2022}, CyberBattleSim \cite{microsoft_defender_research_cyberbattlesim_2021}, and the CybORG environment \cite{standen_cyborg_2021}. These environments are focused on IT networks and currently lack the ability to train multiple defenders. The existing environments do not reflect the challenges of cyber defence on an OT Industrial Control System (ICS). IPMSRL aims to reflect the nuances of OT more accurately in an abstract simulator and provide a platform for defensive agents to be trained to recover an IPMS from a cyber-attack.

We applied Multi-agent Reinforcement Learning (MARL) to the cyber defence of ICSs because of the collaborative nature of the problem. Agents working together distributed across a network taking coordinated remedial actions is likely to be more successful than a single, or multiple uncoordinated, agent(s) working independently.

\section{Operational Technology Scenario}

This paper focused on a generic IPMS, a form of Industrial Control System used onboard ships and submarines.

An IPMS provides the capability for remote monitoring, control and management of the ship’s machinery systems and damage control from key components. This real-time monitoring and control facilitates continuous situational awareness of the machinery state and damage condition of the ship.

IPMS controls and monitors many ship systems across Propulsion, Power, Steering, Stability, Auxiliary and Ancillary systems as well as those providing Damage Control and Fire Fighting (DCFF). To achieve this, IPMS utilises a distributed control system architecture that facilitates interfaces with sensors, equipment, plants, software-based control systems and network-based data. 

An IPMS produces alerts in event of failures, anomalies, or issues. These alerts are not necessarily cyber security alerts; however, they are likely to be instrumental in the context of cyber-attack detection and response. These alerts were assumed to be fed into additional Cyber Threat Detection and Security Information and Event Management (SIEM) tools, leveraging IPMS inputs to detect and generate relevant attack alerts.
The IPMS architecture is based upon traditional Industrial Control System (ICS) design, inclusive of Remote Terminal Units (RTUs), Programmable Logic Controllers (PLCs) and multifunction control consoles providing Human Machine Interfaces (HMIs) combined by a dual redundant network backbone. Specifically, HMI refers to IPMS consoles as well as panels on the equipment. The architecture work has been deliberately abstracted and generalised using open-source information.

\section{IPMSRL - Simulation Environment}

In this paper we introduce IPMSRL, a highly configurable network based multi-agent RL environment. IPMSRL simulates an abstracted generic maritime IPMS where defensive agents attempt to restore an infected network. Meanwhile an attacker attempts to propagate through the network and disrupt IPMS and controlled systems to negatively impact operations.

The IPMSRL network comprises two types of node: critical nodes and non-critical infectable nodes. Critical nodes represent components of core-controlled systems critical to effective operation of the vessel. Two controlled systems are modelled, Chilled Water Plants (CWPs) and the Propulsion System. This selection represents a diverse subset of available controlled systems. Infectable nodes represent HMIs, RTUs, Local Operating Panels (LOPs), PLCs, and network switches. 

\begin{figure}[htb!]
    \centering
    \includegraphics[scale=0.5]{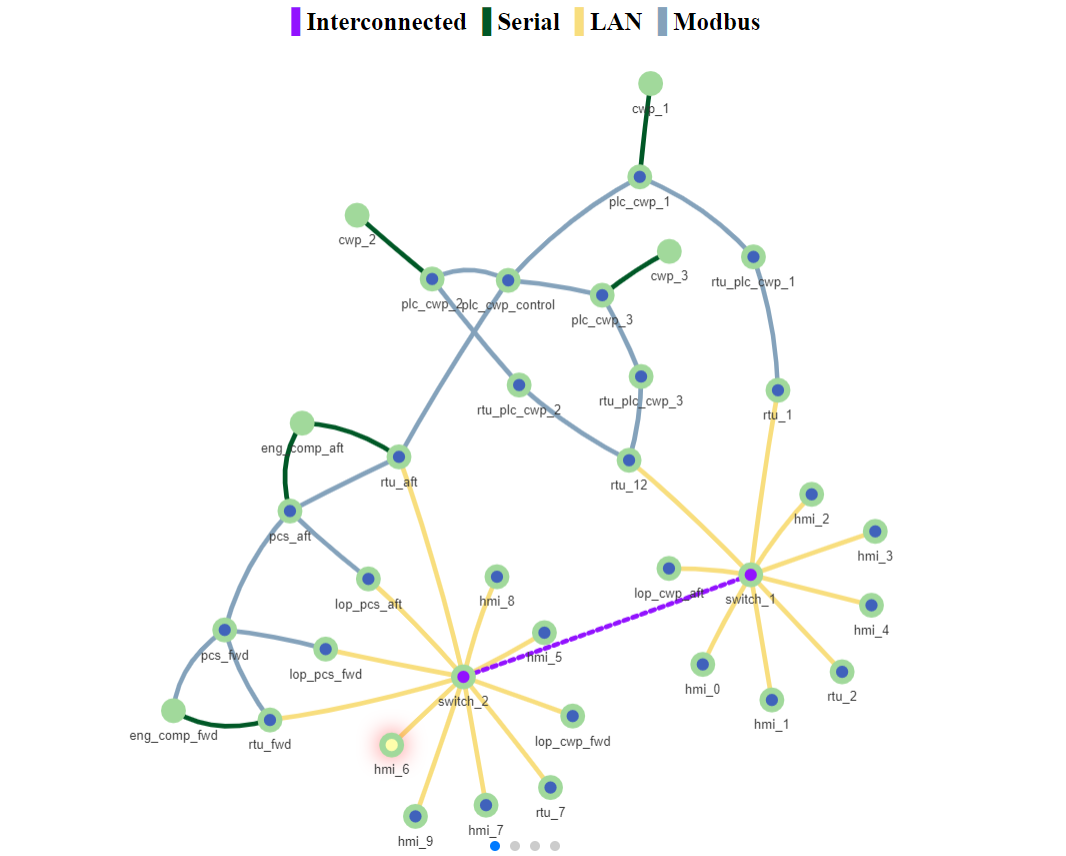}
    \caption{Example of an IPMSRL Network Topology.}
    \label{fig:network}
\end{figure}

The properties of the links between nodes, shown in Figure \ref{fig:network}, differ depending on their type. In Figure \ref{fig:network}, the star representation of the network is for visualisation purposes only. Any HMI/RTU/LOP node which is connected to a network switch is adjacent to any other node directly connected to a network switch, i.e., they are interconnected. The network holds redundancy in its structure through its dual ring network backbone.

\subsection{MITRE ATT\&CK® Framework}

MITRE ATT\&CK\footnote{© 2023 The MITRE Corporation. This work is reproduced and distributed with the permission of The MITRE Corporation.}® ICS \cite{noauthor_mitre_2023}, provides detailed analysis of typical Tactics, Techniques, and Procedures (TTPs) adopted by cyber attackers, which is relevant to industrial control systems. 

Fine-grained attack steps, based on MITRE ATT\&CK\textregistered \space ICS, were implemented to enable more realism, context and complexity to the cyber defensive remedial actions. At different MITRE ATT\&CK\textregistered \space ICS stages the attacker will have different capabilities e.g., lateral movement by leveraging an HMI’s network card and connections. Different defensive remedial actions are required depending on the MITRE ATT\&CK\textregistered \space ICS stage that the attacker has exploited.

The twelve attack stages, represented in IPMSRL, correspond to the associated MITRE ATT\&CK\textregistered \space ICS Tactics are shown in Figure \ref{fig:mitre}:

\begin{figure}[htb!]
    \centering
    \includegraphics[width=1\linewidth]{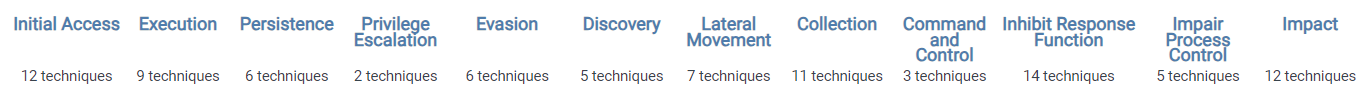}
    \caption{MITRE ATT\&CK\textregistered \space ICS Tactics with the number of techniques for each tactic \cite{noauthor_mitre_2023}.}
    \label{fig:mitre}
\end{figure}

\subsection{Attacker}

The goal of the attacker is to compromise the vessel’s operational capacity, for example, by disrupting IPMS systems and controlled systems (CWPs, Propulsion System) which negatively impacts the vessel.  An attacker propagates through the network by initially infecting a non-critical infectable node, increasing the node’s infection level using an abstraction of the MITRE ATT\&CK\textregistered \space ICS \cite{noauthor_mitre_2023}, framework, and then infecting an adjacent node. Any given node after infection acts independently. It is therefore possible that if multiple nodes are infected then each infected node can progress the infection or laterally move in the same timestep. This makes it very difficult for the defender(s) to recover the network to a clean state once an infection has spread.

Attackers can be created on a sliding scale from fully targeted to fully viral. Fully targeted attackers move directly towards critical nodes displaying ‘knowledge’ of the network. Fully viral attackers move randomly to any adjacent node. Partially viral attackers move towards critical nodes but may also move randomly with a given configured probability.

Attackers also have other configurable attributes such as the probability an infection will progress through the MITRE ATT\&CK\textregistered \space ICS stages or the probability of a successful lateral move.

\subsection{Alerts}

Each of the infectable nodes has an alert system which can be triggered when an infection is present or when an attacker initially tries to infect a non-infected node. The probability that an alert is successfully parsed to the defender(s) allows the user to configure the scale of asymmetric information within the environment. In the context of this paper, an assumption was made that IPMS-generated alerts were fed into a SIEM for further processing and cyber threat detection. The alert success probability, therefore, directly impacts the partial observability of the defender’s observation space.

It is worth noting that the alerts are static in nature. This means that if an alert is set off at a certain timestep, the defender will receive information about the progress of the node infection at a given timestep. No further information about the progression of the infection will be given to the defender unless another alert is set off or the defender takes a remedial action on a node.

\subsection{NIST SP-800-61}

\begin{figure}[htb!]
    \centering
    \includegraphics[width=0.8\linewidth]{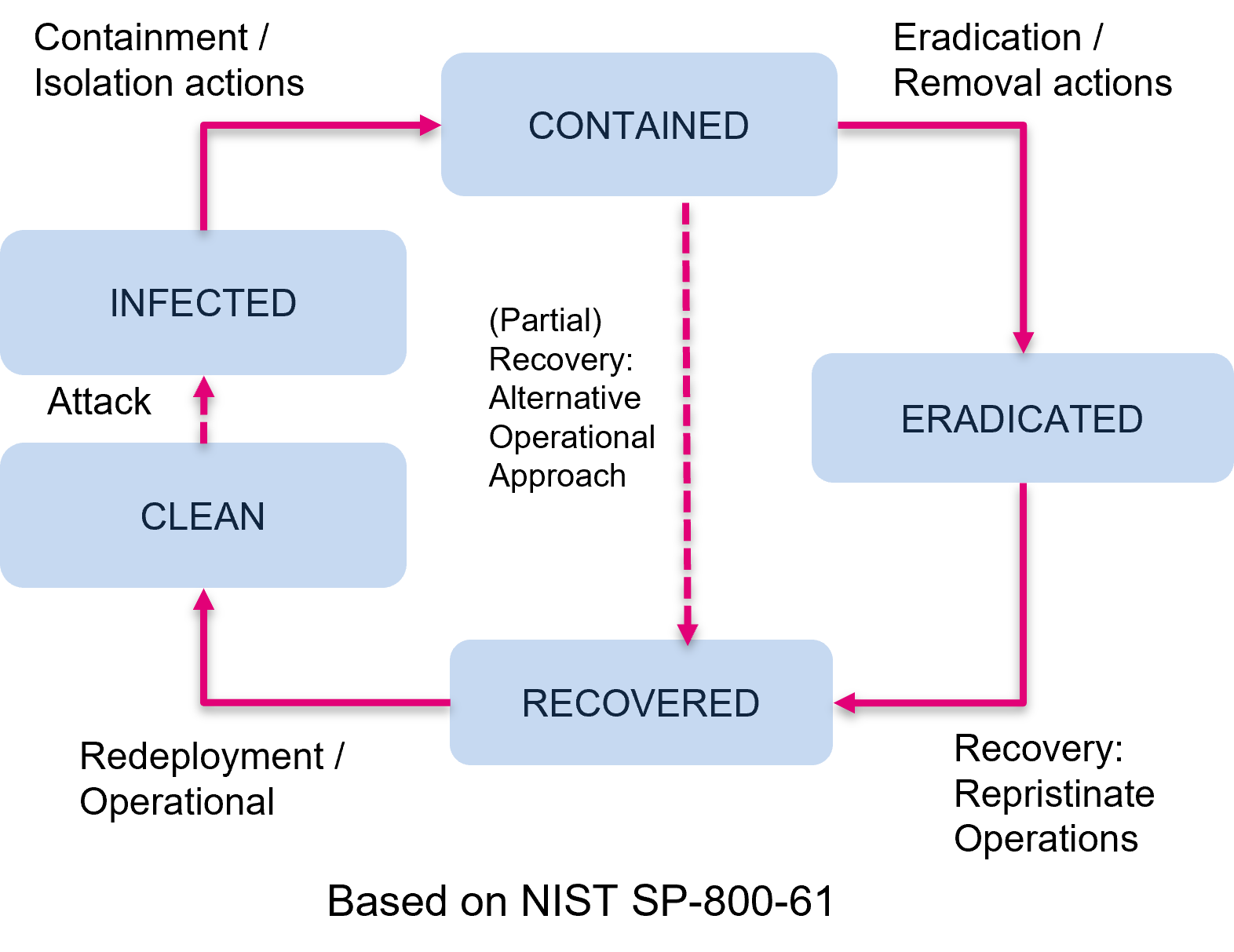}
    \caption{Summary of NIST SP-800-61 Remedial Actions, adapted to targeted Maritime / OT scenario.}
    \label{fig:nist}
\end{figure}

NIST SP-800-61 \cite{cichonski_computer_2012}, describes a standard process to handle cyber incidents and responses. The three key cyber defensive steps relevant to the problem space are Contain, Eradicate, and Recover, shown in Figure \ref{fig:nist}. Each of these types of action need to be taken in a logical order based on cyber defensive best practice. In general, this will be to contain an attack to avoid it spreading across systems, eradicate the attack to remove it from the infected system and recover a node to return the system back to an operational state.
The order of these actions is important and other factors such as the severity of infection impact the effectiveness and ease of taking these actions.

\subsection{Defensive Remedial Actions}

Each defensive agent can take four types of action: Contain, Eradicate, Recover and Wait. These are aligned with the real-world cyber incident response process indicated in NIST SP-800-61 discussed above. 

Containing makes a given node un-operational but prevents the infection from propagating from that node. Eradicating cleans a node of infection. Recovering returns the node to an operational state. For critical nodes, only the recover and wait actions are available whereas, for all other nodes, all actions are available. At each timestep, each defender in succession takes an action. Each of these actions have a configurable delay where a timestep lag is applied depending on the infection state of the node and type of action. This delay tries to reflect the real-world differences between taking simple actions against early infections e.g., modifying credentials, and taking more extreme actions to prevent unrecoverable damage e.g., quarantining systems. Additional dialogue on valuing the costs of remedial actions is considered in later discussion about the reward function. 

IPMSRL supports partial observability in which each defensive agent has its own view of the network state. In addition to the alerts, defensive agents can investigate a node's true infection information by interacting with that node. This knowledge is not shared amongst other defensive agents.

\subsection{Reward Function}

IPMSRL features a highly customisable reward function for both global and intrinsic rewards. Global rewards are awarded to the agents at the end of the episode, whereas intrinsic rewards are awarded within an episode\footnote{The global and intrinsic rewards are divided and shared equally between each agent.}. Global rewards have been subdivided into mission objective and state rewards. The implemented reward function aims to tackle the problem of sparse global rewards, where agents struggle to train since they only receive a small number of reward signals, often at the end of an episode. This is a known problem within RL, and specifically within applied control problems such as robotics \cite{qureshi_intrinsically_2018}. The reward function used in IPMSRL uses a combination of intrinsic rewards and reward shaping \cite{popov_data-efcient_nodate} \cite{mguni_ligs_2022} to allow for further experimentation into how these additions positively/negatively affect training.
The mission objective is defined as follows for win/loss/draw. For episode step $t$ and max steps in an episode $T$, a win, draw or loss is defined as:
\begin{itemize}
    \item Win (+1) – There are no infected nodes or contained nodes, and $t<T$.
    \item Loss (-1) – Any of the critical infrastructure has been infected, and $t<T$.
    \item Draw (+0) – $t = T$ and neither the win nor loss criteria has been met for t steps.
\end{itemize}

For our experiments, $T=50$.\\

The state reward reflects the impact to non-critical and critical nodes during the episode and is graded as low, medium or high. Critical nodes, such as PLCs, RTUs and PCSs that directly control machinery or switches, are more strongly penalised than other nodes, such as HMIs. The weighting of penalties was informed by the failure effects and impact analysis of a generic IPMS. State reward is produced by a 50/50 split of ‘state generic’ and ‘state specific’ rewards:

State Generic – this refers to penalties that are provided when nodes within the environment meet certain conditions e.g., a node is contained but not infected or a node is uninfected and contained without being recovered.

State Specific – this is defined, in levels of severity, as the number of different types of nodes which had been infected in an episode. For example, if two HMIs were infected then the smallest negative reward value (defined in the config) is applied. 

IPMSRL has the ability for agents to receive intrinsic rewards called action score rewards. The defender(s) receive a negative reward depending on the action chosen and the status of the node that is being interacted with (graded as low, medium or high). Nodes with a more severe infection status receive a higher penalty. The agents are therefore incentivised to deal with infections early when they are first alerted. The weighting of the reward function components are user configurable and will be discussed in further detail within the experimental results section.

\section{Multi-Agent Reinforcement Learning (MARL)}

In this paper, two MARL algorithms, IPPO \cite{de_witt_is_2020}, and MAPPO \cite{yu_surprising_2022}, were tested\footnote{Multi-Agent Deep Deterministic Policy Gradient (MADDPG) and Q-MIX were also explored but the implementations within Ray and RLlib were found to be unstable during training on IPMSRL.}. Our implementation of IPPO is equivalent to the IPPO algorithm referenced in the original paper \cite{de_witt_is_2020}. In IPPO, each agent has an actor and critic network with the actor network representing the policy, and the critic network representing the value function. These actor and critic networks are independent to each agent. 
Our MAPPO implementation is slightly different to the MAPPO algorithm used in the original paper \cite{yu_surprising_2022}. The original paper states that their MAPPO algorithm uses shared parameters for both the policy and critic networks. The MAPPO implementation used in this paper uses independent policy networks for each agent, but uses a shared centralised critic also referred to as a centralised value function. An important note is that in other papers, some MAPPO implementations use shared information between agents by using the joint observations and action spaces\footnote{This is the same as the implementation within MARLlib \cite{hu_marllib_2023}, using the global state within training.}. 
Both Proximal Policy Optimisation (PPO) based algorithms were stable during training and demonstrated the ability to recover an infected network within IPMSRL. Two defensive agents were used for all the experiments in this paper. The hardware used for training was a 6 core CPU, 56GB RAM and a NVIDIA Tesla K80 with 12GBs of vRAM. Ray Tune \cite{liaw2018tune} and RLlib were utilised for training. Training the PPO based algorithms for one million timesteps took approximately one hour. The performance metrics used for the experiments were mean episode reward, mean episode length and mean episode outcome which is the mean of the episode outcomes (win (+1)/draw (+0.5)/loss (0)).

\subsection{Hyperparameter Tuning with MAPPO}

\begin{figure}[htb!]
    \centering
    \includegraphics[width=1\linewidth]{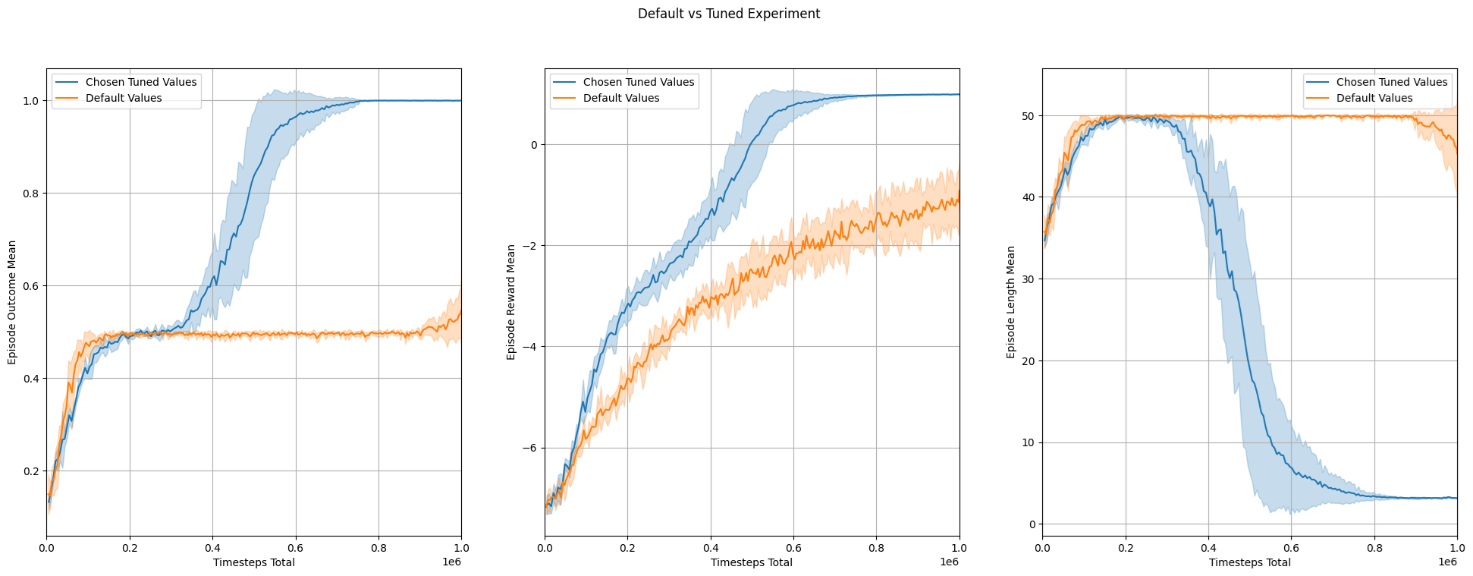}
    \caption{Comparison between default and tuned hyperparameters on MAPPO with 90\% CI.}
    \label{fig:default_tuned}
\end{figure}

An Initial hyperparameter tuning stage to create a baseline for the environment parameter experiments was completed. A brief grid-search over commonly successful hyperparameters was used for eleven PPO specific and general machine learning hyperparameters. In Figure \ref{fig:default_tuned}, a clear difference between the training performance between the tuned and default hyperparameters can be seen. The hyperparameters which were most influential in the improved performance were the Learning Rate, Gamma, Lambda and the Clip Parameter. 
When MAPPO was trained using the default hyperparameters, it converged to an optimal policy more slowly than the tuned model, but it was still able to reach an optimal policy. MAPPO was also used during the Experimental Results section. Throughout the experiments, it was found that in general, PPO based algorithms were still able to improve their performances during training with most configurations of hyperparameters.

\subsection{IPPO vs MAPPO}

\begin{figure}[htb!]
    \centering
    \includegraphics[width=1\linewidth]{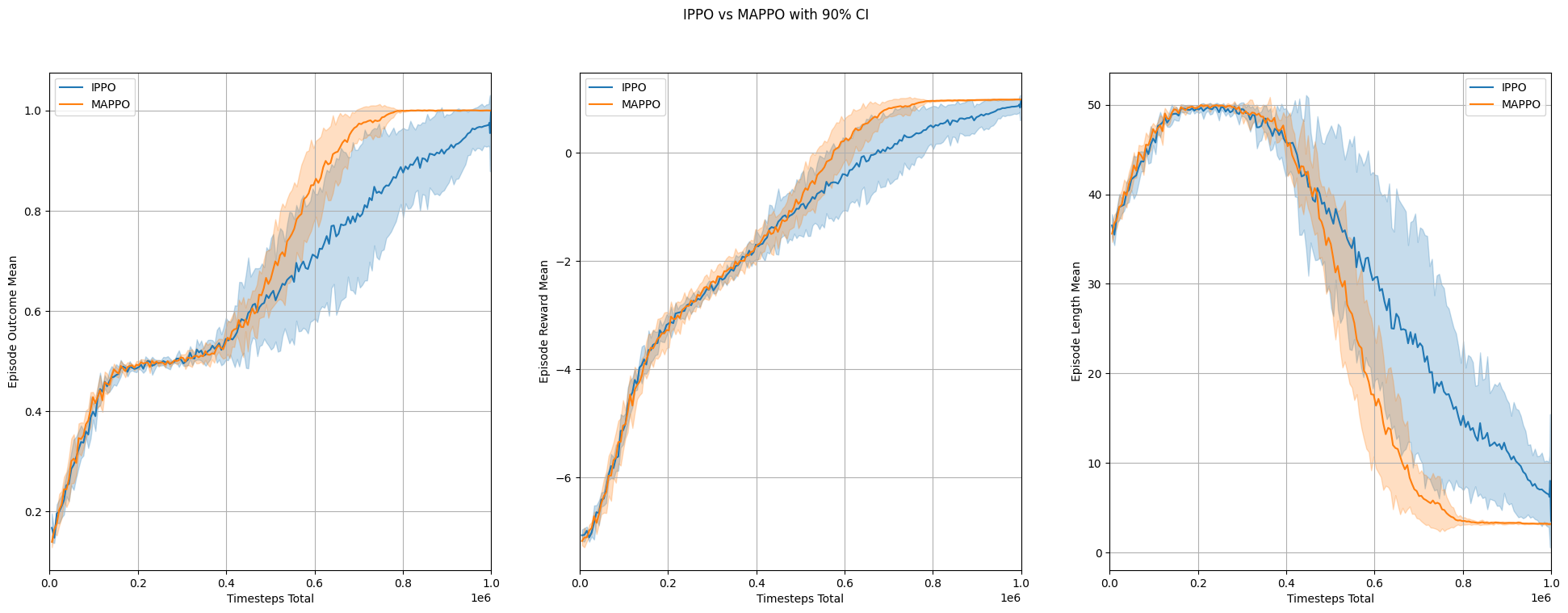}
    \caption{IPPO vs MAPPO Experiment with a CI of 90\%}
    \label{fig:ippo_mappo}
\end{figure}

MAPPO outperformed IPPO, converging to the optimal policy quicker. This statement is based on the higher win rate, higher reward total and the ability to win an episode in fewer episode steps, suggesting a more efficient strategy has developed. The point at which the two algorithms diverge in Figure \ref{fig:ippo_mappo} underpins some of the differences in performance. At 200K timesteps IPPO and MAPPO have similar performance and are drawing most episodes. After this point, both IPPO and MAPPO start to develop a winning policy but with MAPPO optimising its policy at a faster rate. In our experimentation, MAPPO clearly gains an advantage using the same critic network. The centralised critic allows for the agents to find a better policy where agents are collaborative faster than without the shared critic network. This is the converse to what has been found elsewhere in the literature, where the addition of a shared value function reduced the performance significantly \cite{de_witt_is_2020}.

MAPPO's confidence interval (90\%) bands were much narrower than IPPO. The narrower area of the confidence interval suggests that the individual trials of MAPPO algorithm were more stable than IPPO, implying the training is more consistent. This experiment demonstrates that the centralised critic can help improve the policy updates at each training epoch.

\newpage

\section{Experimental Results}

\subsection{Alert Success Probability}

A configurable attacker parameter is the alert success probability: this is the probability of an alert being set off on any infected node after an infection on a node progresses successfully laterally to another node.

\begin{figure}[htb!]
    \centering
    \includegraphics[width=1\linewidth]{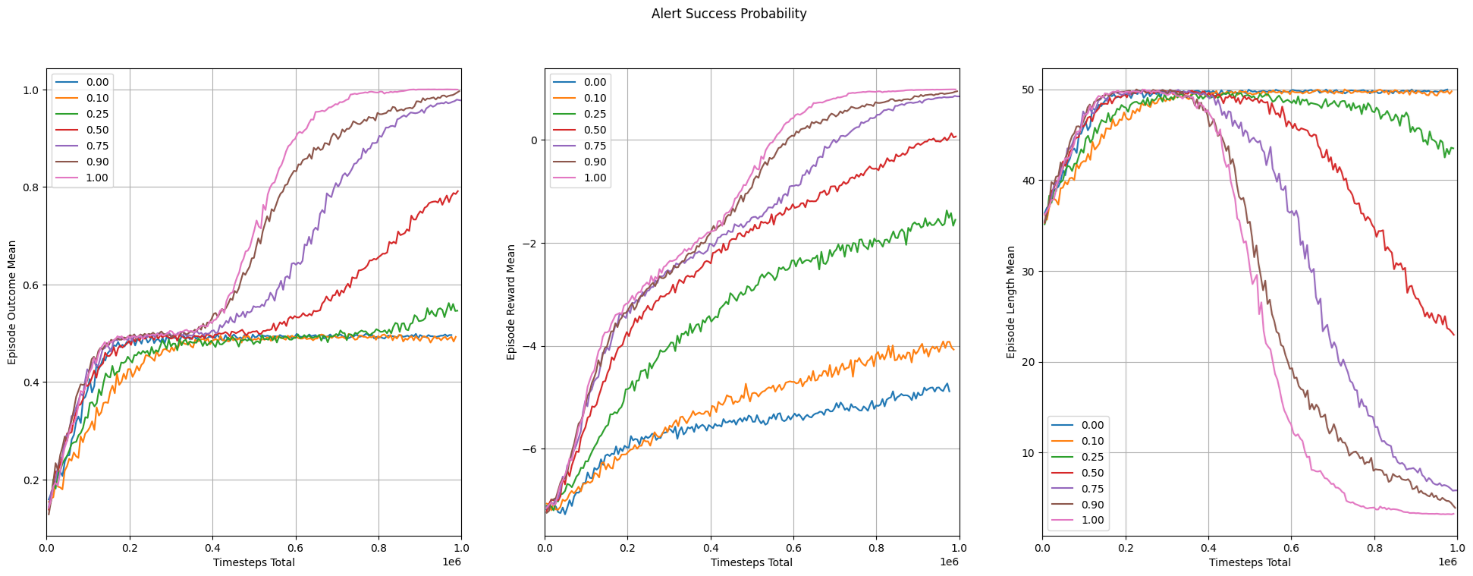}
    \caption{Alert Success Probability Experiment.}
    \label{fig:alert_success}
\end{figure}

Figure \ref{fig:alert_success}.1 shows that the episode outcome mean for probabilities 0.00 (blue plot) and 0.1 (orange plot) converges to 0.5. This suggests that almost every episode is ending in a draw. Figure \ref{fig:alert_success}.3 shows they all converge to episode length mean of 50. This implies the defender is not receiving enough information to win but is managing to draw nearly all episodes. 

In Table \ref{table:alert_success_results}, the impact of the defenders receiving more accurate information about the environment is clear. The more accurate the information they receive, the better they can perform. Without an alert success probability of 1, within the training time set, none of the parameters were able to produce an optimal strategy. However, an agent with an alert probability of 0.75 or 0.9 was still able to win in over 97.5\% or 99.5\% of episodes, respectively. These results suggest that the agents can learn effective strategies in partially observable environments, but that the SIEMs alert detection and processing is important to the success of autonomous cyber defence.

\begin{table}[htb!]
    \centering
    \caption{Alert Success Probability Experiment Results.}
    \begin{tabular}{
    >{\columncolor[HTML]{DCE2DF}}l 
    >{\columncolor[HTML]{DCE2DF}}l 
    >{\columncolor[HTML]{DCE2DF}}l 
    >{\columncolor[HTML]{DCE2DF}}l l}
    \cellcolor[HTML]{005581}{\color[HTML]{FFFFFF} Parameter} & \cellcolor[HTML]{005581}{\color[HTML]{FFFFFF} Episode    Outcome Mean} & \cellcolor[HTML]{005581}{\color[HTML]{FFFFFF} Episode    Reward Mean} & \cellcolor[HTML]{005581}{\color[HTML]{FFFFFF} Episode    Length Mean} &  \\
    0                                                        & 0.4963                                                                 & -4.8858                                                               & 49.9803                                                               &  \\
    0.1                                                      & 0.4927                                                                 & -4.0741                                                               & 49.7891                                                               &  \\
    0.25                                                     & 0.5463                                                                 & -1.5508                                                               & 43.5021                                                               &  \\
    0.5                                                      & 0.7913                                                                 & 0.0672                                                                & 22.9676                                                               &  \\
    0.75                                                     & 0.9772                                                                 & 0.845                                                                 & 5.8354                                                                &  \\
    0.9                                                      & 0.9962                                                                 & 0.9454                                                                & 3.9113                                                                &  \\
    1                                                        & 0.9995                                                                 & 0.9801                                                                & 3.2588                                                                & 
    \end{tabular}
    \label{table:alert_success_results}
\end{table}

\subsection{Reward Experiments}

\begin{figure}[htb!]
    \centering
    \includegraphics[width=1\linewidth]{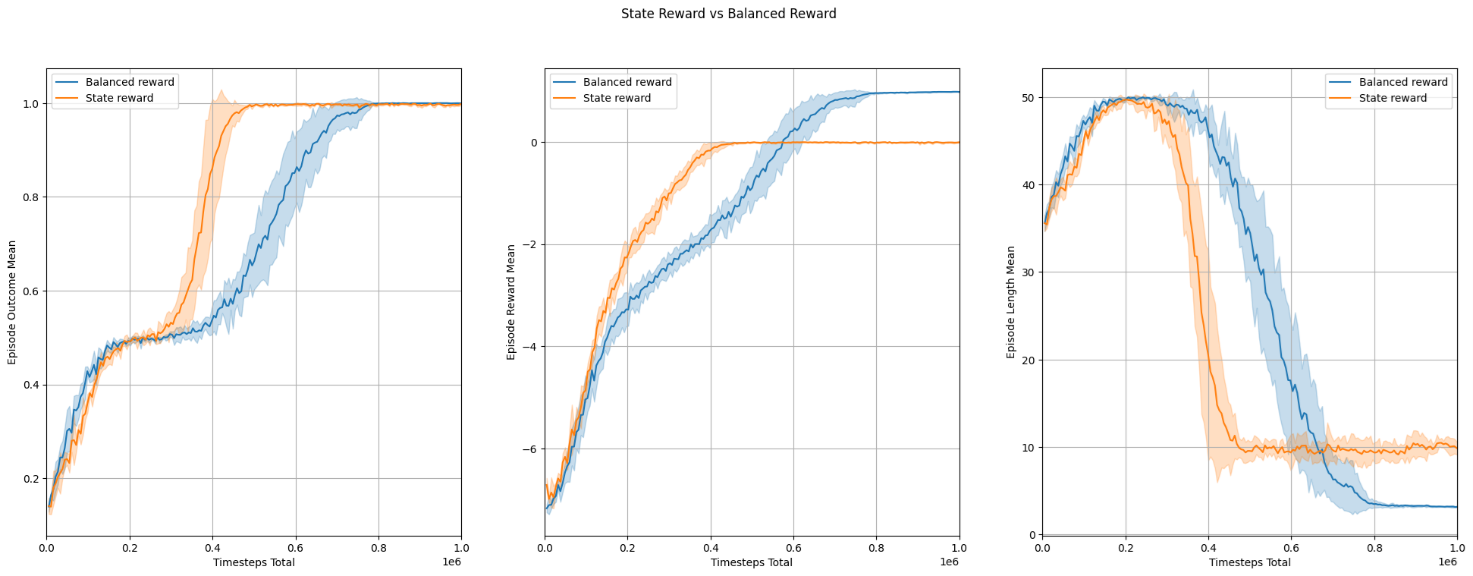}
    \caption{Reward Shaping Experiment - State Reward vs Balanced Reward with 90\% CI.}
    \label{fig:reward_shaping}
\end{figure}

In Figure \ref{fig:reward_shaping}, an investigation into the difference in training performance between state-based rewards and balanced rewards was conducted. State based reward refers to a reward function solely made up of the state rewards discussed in the reward function section. Whereas balanced reward also includes intrinsic rewards, given to the agents during an episode based on their actions, and an overall score based on the episode outcome. Using solely state rewards, convergence towards a winning strategy was faster than for balanced rewards, but performed less optimally overall than balanced rewards. Figure \ref{fig:reward_shaping} shows the two parameter sets plotted with 90\% confidence intervals, validating their significant differences in behaviour.

Figure \ref{fig:reward_shaping}.1 shows that once the state-based reward parameter reaches an episode outcome mean of almost 1, indicating it has won almost every episode, the episode length mean, shown in Figure \ref{fig:reward_shaping}.3 reaches a value around 10 steps per episode. It then plateaus for the duration of the training. This is a far higher mean steps per episode than the balanced reward agents. A lower episode length mean suggests that the defender(s), if the win rate is near 100\%, find a more efficient defensive strategy. The balanced reward approach, which initially converges toward a winning strategy more slowly, reaches an episode length mean of 3.18. This is a dramatic improvement and represents a far more effective and performant strategy.
The state reward agents have no incentive to improve or fine-tune their behaviour further than the condition required to reliably win. The agents trained with just state reward are therefore 'lazy' with their action choices, taking more actions than is necessary to win an episode. 
This experiment shows the value of reward shaping and how important the dynamics of agent incentives are to a performant outcome.  

\section{Conclusion and Future Work}

This paper introduced IPMSRL, a highly configurable network based multi-agent RL environment and demonstrated the capabilities of MARL agents to successfully recover an abstract IPMS to an operational state following a cyber-attack. Findings demonstrate that hyperparameter tuning, reward shaping and the quality of alerts are all important aspects in developing performant agents.

A key issue with using a simulator to train an RL agent is the sim-to-real-gap. This describes the differences present between real systems and the simulated environments that are used to replicate them and the problems that even small differences can cause. The issues are compounded further when environments are intentionally abstracted away from the real system for reasons of dimensionality, security or complexity. RL simulators such as IPMSRL are important for developing the core concepts needed to build confidence and understanding in the techniques being employed, such as MARL. But more realistic simulators or emulators will be required to facilitate movement towards applying these concepts on real systems, which should be the eventual goal.

Another avenue of future research is the generalisability of the agents. Agents will need to be able to adapt to different types of attack, scenario (where components may be valued differently) and network topology. Without demonstrating this flexibility, agents are unlikely to be stable enough to be trusted in real world control systems.

\begin{acknowledgments}
Research funded by Frazer-Nash Consultancy Ltd. on behalf of the Defence Science and Technology Laboratory (Dstl) which is an executive agency of the UK Ministry of Defence providing world class expertise and delivering cutting-edge science and technology for the benefit of the nation and allies. The research supports the Autonomous Resilient Cyber Defence (ARCD) project within the Dstl Cyber Defence Enhancement programme.

The authors would also like to thank Clare Jubb, Andy Pollard, Christos Giachritsis, Jake Rigby, Ben Golding, Julie Kimbrey and Brian Bassil for their wider contribution to the project and paper.
\end{acknowledgments}



\end{document}